\documentclass{article}

\usepackage{arxiv_no_preprint}

\usepackage[utf8]{inputenc} 
\usepackage[T1]{fontenc}    
\usepackage{hyperref}       
\usepackage{url}            
\usepackage{booktabs}       
\usepackage{amsfonts}       
\usepackage{nicefrac}       
\usepackage{microtype}      

\usepackage[dvipsnames, table]{xcolor}  
\usepackage{mathtools}

\usepackage{setspace}
\usepackage{mdwlist}
\usepackage[title,titletoc,toc]{appendix}

\usepackage{times}
\usepackage{epsfig}
\usepackage{graphicx}
\usepackage{amsmath, amsthm, amssymb}
\usepackage{mdwlist}

\usepackage{enumitem}
\usepackage{siunitx}

\usepackage[ruled,vlined]{algorithm2e}

\usepackage{pslatex}           

\usepackage{epsfig}

\usepackage{amsfonts}
\usepackage{amsmath, amsthm, amssymb}

\usepackage[ruled,vlined]{algorithm2e}

\usepackage{caption}
\usepackage[nocompress]{cite}
\usepackage{amsmath}
\usepackage{float}
\usepackage[table]{xcolor}
\usepackage{array}
\usepackage{url}
\usepackage{boldline}
\usepackage{listings}

\makeatletter
\def\ps@IEEEtitlepagestyle{%
  \def\@oddfoot{\mycopyrightnotice}%
  \def\@evenfoot{}%
}
\def\mycopyrightnotice{
{ \footnotesize
  \begin{minipage}{\textwidth}
  \centering
  \copyright2016 IEEE. Personal use of this material is permitted. Permission from IEEE must be obtained for all other uses, in any current or future media, including reprinting /republishing this material for advertising or promotional purposes, creating new collective works, for resale or redistribution to servers or lists, or reuse of any copyrighted component of this work in other works. \hfill
\end{minipage}
}
}

\usepackage{amsmath} 
\usepackage{amssymb}  
\usepackage{verbatim}  
\usepackage{graphicx}
\usepackage{subcaption}
\usepackage{array}

\usepackage{float}
\usepackage{placeins}
\usepackage{hhline}
\usepackage{multirow}

\title{4-D Scene Alignment in Surveillance Video}

\author{
  Robert Wagner, Daniel Crispell, Patrick Feeney, Joe Mundy\\
  Vision Systems, Inc.\\
  Riverside, RI 02915\\
  \texttt{\{robert.wagner, dan, patrick.feeney, jlm\}@visionsystemsinc.com}\\
}

\begin{document}
\maketitle

\begin{abstract}
Designing robust activity detectors for fixed camera surveillance video requires knowledge of the 3-D scene.  This paper presents an automatic camera calibration process that provides a mechanism to reason about the spatial proximity between objects at different times.  It combines a CNN-based camera pose estimator with a vertical scale provided by pedestrian observations to establish the 4-D scene geometry.  Unlike some previous methods, the people do not need to be tracked nor do the head and feet need to be explicitly detected.  It is robust to individual height variations and camera parameter estimation errors.
\end{abstract}


\section{Introduction}
Automatic activity detection is a extremely challenging task, especially when the image fidelity is low.  In fact, the 2018 ActEv challenge program reported a top weighted probability of $0.750$ missed detections at 0.15 false alarms per minute averaged over activity types in phase one, which is considered unusable for system development.  The ActEv challenge is based on the VIRAT dataset and the activities of interest include people interacting with vehicles, e.g.  {\textit{Open Trunk}, \textit{Closing Trunk}, \textit{Exiting}, and \textit{Entering}}. There many instances where the desired activities are in the far field and the images of people are tiny, i.e. less than 32px.

One design consideration for developing a system to detect novel activities, including people interacting with vehicles, is to use natural language.  One can imagine a surveillance system where a user could provide a verbal description of the activity in question.  The natural language description of an activity can be decomposed into a set of semantic predicates.  These predicates provide a recipe for identifying patterns in the video corpus.  For example, say the user provides the input ``person enters vehicle.''  3-D geometric constraints can be used to decompose this activity description into meaningful space-time predicates such as:
\begin{itemize}[noitemsep]
\item \it{person} \textbf{moving towards} \it{vehicle}
\item \it{person} \textbf{near} \it{vehicle}
\item \it{person} \textbf{disappears} from view
\end{itemize}

The scene geometry needs to be estimated so that a metric can be imposed that allows distance measurements in 3-D.  The main advantage of this approach is that actions in 4-D space-time are viewpoint invariant -- meaning when a person is near a vehicle, they will be considered ''near the vehicle'' from any observation point where both objects are visible.  Estimating the 3-D positions of objects will also greatly improve the chances of detecting activities of interest, even in the case of tiny people.

The 4-D scene alignment can be accomplished by developing an automated camera calibration methodology.  To start, the method described in this article requires an object of known size to be imaged.  For video surveillance applications, it is a fair assumption that people will appear in the scene at some point.  Therefore, the method presented makes use of person detections as well as a convolutional neural network (CNN) that estimates the horizon line and vertical vanishing point from a single scene image.  

\begin{figure}[h!]
\centering
\includegraphics[height=2in]{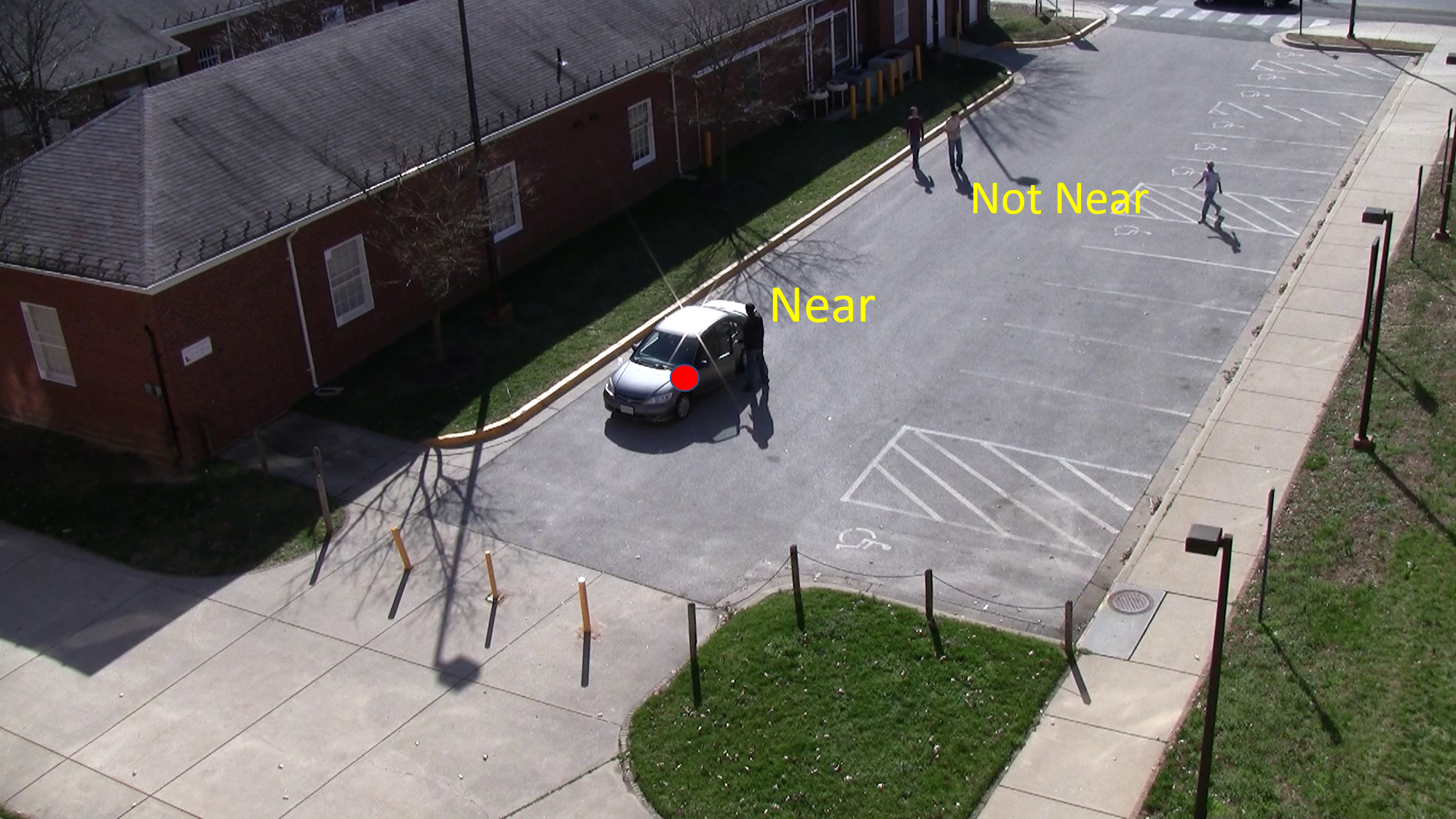}
\caption{Persons within a specified distance to the centroid of vehicle (indicated by the red dot) are considered near.  Distances are measured on the ground plane.}
\label{f:p_near_example}
\end{figure}

The calibration method provides the necessary scene geometry that facilitates development of a working definition of the space-time predicate \textit{person} \textbf{near} \textit{vehicle}, as illustrated in Figure \ref{f:p_near_example}.

The remainder of the paper is laid out as follows. A selection of related prior works is presented in Section \ref{sec:relatedWork}. The details of the proposed method for 3-D location estimation for pedestrians and vehicles is described in Section \ref{sec:method}. Quantitative results produced by the method are presented in Section \ref{sec:results}. Finally, the paper is concluded in Section \ref{sec:conclusion}.

\section{Related Work}
\label{sec:relatedWork}

Estimating the height of objects in images is closely related to 3-D location estimation. As is described in Section \ref{sec:method}, height estimations can be combined with camera parameters to establish a projection matrix from 3-D world coordinates to the 2-D image plane. Vester \cite{vester} provides an overview and analysis of common techniques for height estimation. The first algorithm presented utilizes two user-defined parallel planes to determine an object's height. The second algorithm presented utilizes knowledge of the camera's position and orientation relative to the ground plane as well as a known object height to perform a back projection to 3-D coordinates. Although the formulation slightly differs, this process is most similar to the process used in this paper, with the caveat that this paper estimates the camera's position and orientation and object heights. The final algorithm presented involves setting up 5 reference objects with known heights, one at each corner of the picture and one in the center of the picture. These heights are then used to calculate the height of an unknown object.

Limited work has been done on 3-D location estimation for vehicles from 2-D images. Dubska et al. \cite{dubska} use the assumption that vehicles are moving only forwards or backwards along a straight road to construct 3-D bounding boxes for vehicles. 3-D distance and speed measurements are then made by approximating the real world scale of the bounding boxes from the bounding box of a car of median length, width, and height.

Limited work has also been done for 3-D location estimation for pedestrians from 2-D images. Bertozzi et al. \cite{bertozzi} present a method that relies on two cameras for stereo refinement of bounding boxes before estimation.

Camera pose estimation for a static camera has been thoroughly researched \cite{hollander,dubska,hold-geoffroy,li,liu,shalnov,workmanHorizon,workmanDeepFocal,zhai}. Abbas and Zisserman developed the Bird’s Eye View Network \cite{abbas_birds_eye} for estimation of focal length, camera pitch, and camera roll using image data without specific features. Other methods for featureless estimation of camera pose exist, but do not have code publicly available \cite{hold-geoffroy,li}. There are also featureless convolutional neural networks that focus only on estimating the horizon line \cite{workmanHorizon,zhai} or focal length \cite{workmanDeepFocal} of the camera. Feature-based methods for camera pose estimation have also been created. Methods using pedestrian data as features to predict camera pose exist \cite{hollander,liu,shalnov}, but do not have code publicly available. The aforementioned paper on 3-D location estimation for vehicles \cite{dubska} also predicts camera pose from vehicle data, but requires the previously mentioned assumptions.

\section{Method}
\label{sec:method}

This section describes a method to estimate the $3\times4$ projection matrix using the camera parameters provided by the Bird's Eye View network (i.e. focal length (pixels), camera tilt (degrees), and camera roll (degrees)) and the person detections from a state-of-the-art detector such as Mask R-CNN \cite{He_2017} or manual annotations.  

To establish the vertical scale, one can specify an average height for a person (e.g. \SI{1.778}{\metre}) and use the top center and the bottom center of a person's bounding box as a height estimate of the person in pixels.

The procedure outlined below recovers the camera height, the 3-D locations of the observed people, and the projection matrix.

\subsection{Projection Matrix Estimation}
\label{sec:p_matrix}
The projection matrix $P$ takes a 3-D point in world coordinates and projects it onto the 2-D image plane 
\begin{equation}
\left[
\begin{array}{c}
\lambda u \\
\lambda v \\
\lambda \\
\end{array}
\right]
=
P
\left[
\begin{array}{c}
X \\
Y \\
Z \\
1 \\
\end{array}
\right]
\end{equation}
up to an unknown scaling factor $\lambda$.

The projection matrix is comprised of the intrinsic camera parameters $K$ and the extrinsic camera parameters $[R|t]$ and is typically expressed as:
\begin{equation}
P = K[R|t]
\end{equation}
where the intrinsic parameters consist of the focal length $f$ and the principal point $(u_o, v_o)$:
\begin{equation}
K =
\left[
\begin{array}{ccc}
f & 0 & u_o \\
0 & f & v_o \\
0 & 0 & 1 
\end{array}
\right]
\end{equation}
The camera center can be obtained with:
\begin{equation}
C_o = -R^Tt
\end{equation}
or equivalently :
\begin{equation}
t = -RC_o
\end{equation}
If the camera center is defined as
\begin{equation}
C_o = \left[\begin{array}{c}
0 \\
0 \\
C_Z
\end{array}
\right]
\end{equation}
then, 
\begin{equation}
t =
\left[
\begin{array}{c}
-r_{02}C_Z \\
-r_{12}C_Z \\
-r_{22}C_Z \\
\end{array}
\right]
\end{equation}
and the full expansion of the matrices will be:
\begin{equation}
P =
\left[
\begin{array}{ccc}
f & 0 & u_o \\
0 & f & v_o \\
0 & 0 & 1 
\end{array}
\right]
\left[
\begin{array}{cccc}
r_{00} & r_{01} & r_{02} & -r_{02}C_Z \\
r_{10} & r_{11} & r_{12} & -r_{12}C_Z \\
r_{20} & r_{21} & r_{22} & -r_{22}C_Z \\
\end{array}
\right]
\end{equation}

After matrix multiplication
\begin{equation}
P =
\left[
\begin{array}{cccc}
f r_{00} + u_o r_{20} & f r_{01} + u_o r_{21} & f r_{02} + u_o r_{22} & -f r_{02}C_Z -u_o r_{22}C_Z\\
f r_{10} + v_o r_{20} & f r_{11} + v_o r_{21} & f r_{12} + v_o r_{22} & -f r_{12}C_Z -v_o r_{22}C_Z\\
r_{20} & r_{21} & r_{22} & -r_{22}C_Z \\
\end{array}
\right]
\end{equation}
The correspondence between 3-D world coordinates and 2-D image coordinates can now be rewritten as:
\begin{equation}
\left[
\begin{array}{c}
\lambda u \\
\lambda v \\
\lambda \\
\end{array}
\right]
=
\left[
\begin{array}{cccc}
a & b & c & d C_Z \\
e & f & g & h C_Z \\
i & j & k & l C_Z \\
\end{array}
\right]
\left[
\begin{array}{c}
X \\
Y \\
Z \\
1 \\
\end{array}
\right]
\end{equation}
Performing the matrix multiplication provides the following equations for image points:
\begin{align*}
u =
\frac{aX + bY + cZ + dC_Z}{iX + jY + kZ + l C_Z} \\
v =
\frac{eX + fY + gZ + hC_Z}{iX + jY + kZ + l C_Z}
\end{align*}
These equations can be rearranged as:
\begin{align*}
C_Z(d - u l) + X(a - u i) + Y(b - u j) = u k Z - c Z \\
C_Z(h - v l) + X(e - v i) + Y(f - v j) = v k Z - g Z \\
\end{align*}

Given the intrinsic matrix $K$ and the rotation matrix $R$, a linear system can be set up to solve for the unknown camera height, $C_Z$, and the 3-D locations of the observed people:
\begin{equation}
Ax = b
\end{equation}
where
\begin{align*}
A=
\left[
\begin{array}{cccccccccc}
(d - u_0^g l) & (a - u_0^g i) & (b - u_0^g j) & 0 & 0 & 0 & \hdots & 0 & 0 \\[2pt]
(h - v_0^g l) & (e - v_0^g i) & (f - v_0^g j) & 0 & 0 & 0 & \hdots & 0 & 0 \\[2pt]
(d - u_0^t l) & 0 & 0 & (a - u_0^t i) & (b - u_0^t j) & 0 & \hdots & 0 & 0 \\[2pt]
(h - v_0^t l) & 0 & 0 & (e - v_0^t i) & (f - v_0^t j) & 0 & \hdots & 0 & 0 \\[2pt]
\vdots & \vdots & \vdots & \ddots & \ddots & \ddots & \ddots & \vdots & \vdots\\[2pt]
(d - u_{N-1}^g l) & 0 & 0 & \hdots & 0 & (a - u_{N-1}^g i) & (b - u_{N-1}^g j) & 0 & 0\\[2pt]
(h - v_{N-1}^g l) & 0 & 0 & \hdots & 0 & (e - v_{N-1}^g i) & (f - v_{N-1}^g j) & 0 & 0\\[2pt]
(d - u_{N-1}^t l) & 0 & 0 & \hdots & 0 & 0 & 0 & (a - u_{N-1}^t i) & (b - u_{N-1}^t j) \\[2pt]
(h - v_{N-1}^t l) & 0 & 0 & \hdots & 0 & 0 & 0 & (e - v_{N-1}^t i) & (f - v_{N-1}^t j) \\[2pt]
\end{array}
\right]
\end{align*}

\begin{align*}
b = 
\left[
\begin{array}{c}
Z_{0}^g(u_0^g k - c) \\[2pt]
Z_{0}^g(v_0^g k - g) \\[2pt]
Z_{avg}^t(u_0^t k - c) \\[2pt]
Z_{avg}^t(v_0^t k - g) \\[2pt]
\vdots \\[2pt]
Z_{0}^g(u_{N-1}^g k - c) \\[2pt]
Z_{0}^g(v_{N-1}^g k - g) \\[2pt]
Z_{avg}^t(u_{N-1}^t k - c) \\[2pt]
Z_{avg}^t(v_{N-1}^t k - g) \\[2pt]
\end{array}
\right]
\end{align*}

and $x$ will be in the form $[C_Z, X_0^g, Y_0^g, X_0^t, Y_0^t, ..., X_{N-1}^g, Y_{N-1}^g, X_{N-1}^t, Y_{N-1}^t]$.  In this notation, $(X^g, Y^g, Z_{0}^g)$ are the world coordinates of the person's footprint, $(X^t, Y^t, Z_{avg}^t)$ are the world coordinates of the top of the person's head, and $Z_{avg}^t$ is the average person height in meters.

\begin{figure}[h!]
\centering
\begin{subfigure}[h]{0.45\textwidth}
\centering
\includegraphics[height=2in]{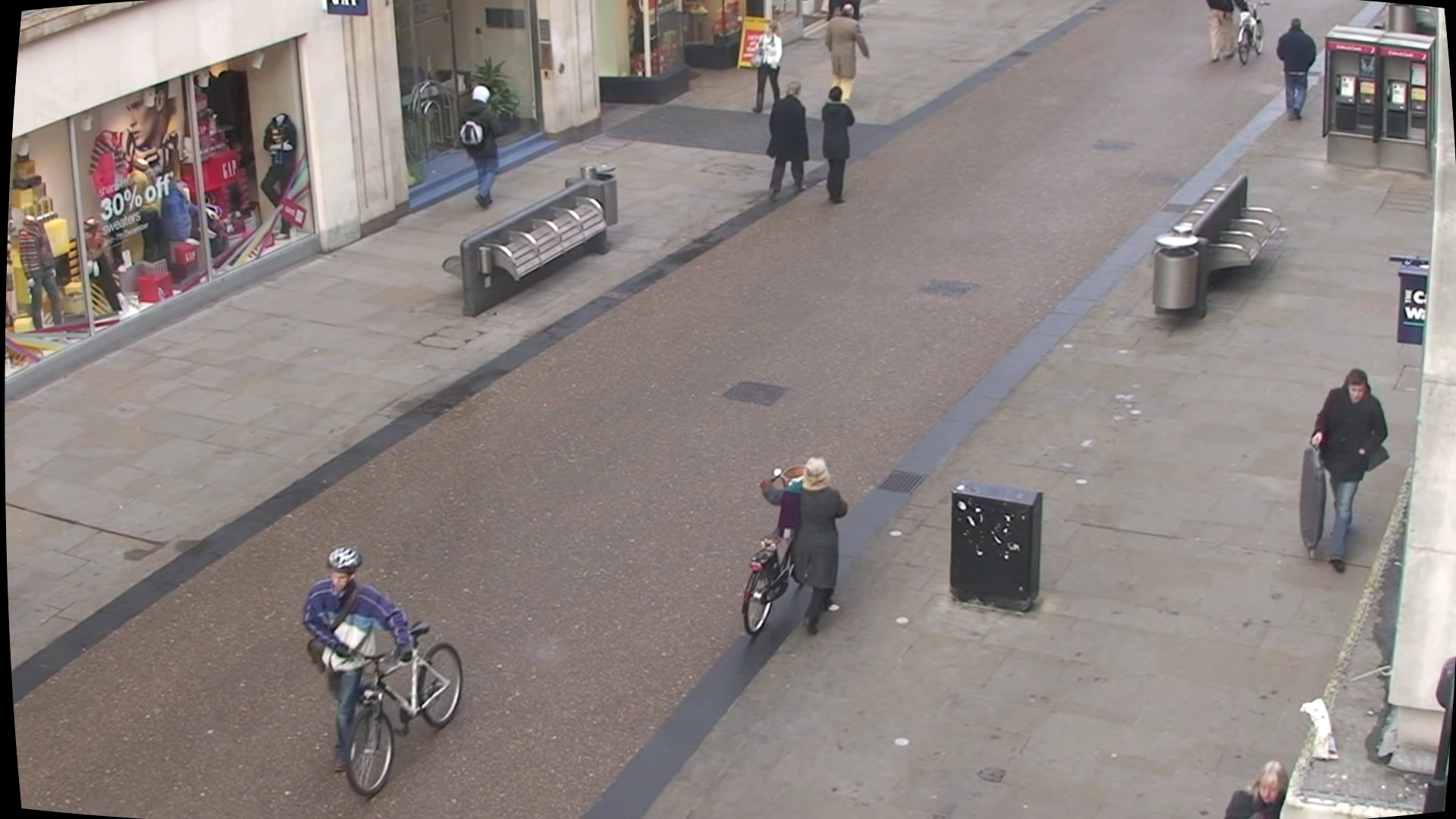}
\caption{Sample image from TownCentre dataset}
\label{f:town_centre_image}
\end{subfigure}%
\hfill
\begin{subfigure}[h]{0.45\textwidth}
\centering
\includegraphics[height=2in]{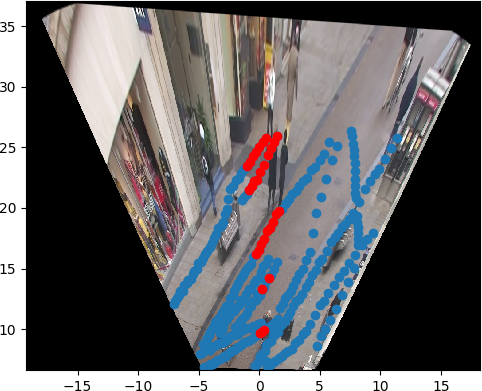}
\caption{RANSAC fit}
\label{f:ransac_fit_to_gt_tracks}
\end{subfigure}
\caption{TownCentre dataset}
\label{f:projection_matrix_ft}
\end{figure}

The top center pixel of a person's bounding box may not correspond to the actual position to the top of the head.  Also, the assumption that all people are the same height is false.  Therefore, RANSAC \cite{fischler} is used to robustly estimate the projection matrix.  The fit is determined by minimizing the reprojection error from the world coordinate estimate of the top of the head position to the pixel location in the image.  Example results are shown in Figure \ref{f:ransac_fit_to_gt_tracks}, where the dots represent the footprints of a person detection.  The red dots are the inliers chosen by RANSAC.  

The number of inliers is small due to specifying a tight tolerance on the reprojection error (i.e. < 5 pixels).  More notably, the inliers tend to cluster in a region where the projected vertical direction is aligned with the vertical image axis, since heights are measured using the vertical dimension of axis-aligned image bounding boxes.

\section{Results}
\label{sec:results}

\subsection{Projection Matrices}
\label{sec:projection_matrices}
\begin{figure}[h!]
\centering
\includegraphics[height=2in]{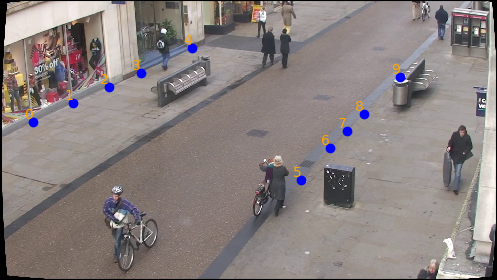}
\caption{TownCentre image correspondence points}
\label{f:img_correspondence_pts}
\end{figure}

The TownCentre dataset was used to evaluate this approach as it provided calibration parameters that were taken as ground truth \cite{Benfold}.  Additionally, a new set of manual annotations was constructed for several people over thousands of frames and a set of correspondance points was defined (see Figure \ref{f:img_correspondence_pts}).

One can establish the effectiveness of the automatic calibration procedure by focusing on the space-time predicate of \textit{person near vehicle}.  Since there are no vehicles in the TownCentre dataset, the image correspondence points can be re-imagined as the centroids of vehicles.  The ground truth projection matrix was used to project the vehicle centroids into the 3-D world.  Similarly, the bottom center of the ground truth person bounding boxes were projected into the 3-D world to represent the footprint of the person in 3-D coordinates.  These points served as the reference points to test the accuracy of the automated calibration procedure.

\begin{figure}[h!]
\begin{subfigure}[h]{0.45\textwidth}
\centering
\includegraphics[height=2.75in]{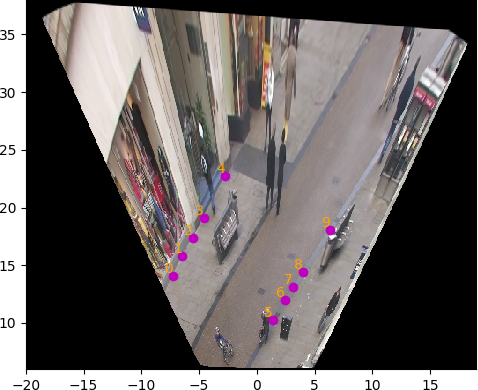}
\caption{Automatic calibration ground plane homography with correspondence points}
\label{f:auto_correspondence_pts}
\end{subfigure}%
\hfill
\begin{subfigure}[h]{0.45\textwidth}
\centering
\includegraphics[width=2.75in]{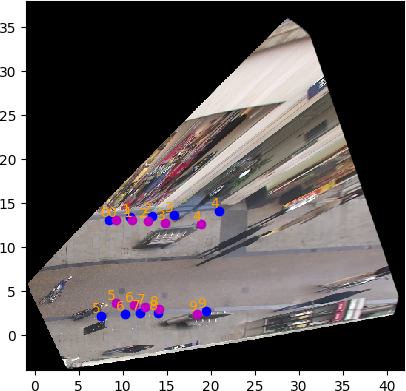}
\caption{Ground truth ground plane homography with ground truth correspondence points in blue and estimated points in magenta}
\label{f:gt_correspondence_pts}
\end{subfigure}
\caption{Ground plane correspondence points}
\label{f:ransac_fit_to_people}
\end{figure}

Prior to running the automatic calibration procedure, radial and tangential distortion was removed from all TownCentre images using the camera calibration parameters included with the dataset.  The Bird's Eye network was run on 1000 images and the mode of the distribution of predicted values were chosen to represent the focal length, camera tilt and camera roll.  The network predicted a wider field of view than is represented in the image. This is most likely a result of the network being trained with the synthetic CARLA dataset that contains images with larger fields of view (FOV).  An error in the FOV estimate will reduce the accuracy of the world distance measurements.

The focal length obtained from the Bird's Eye network can be used to construct the camera intrinsic parameter matrix $K$. The Bird's Eye network also provides the tilt and roll parameters that are used to establish the camera rotation matrix 

\begin{equation}
R = R_{roll} \times R_{pitch} \times R_{overhead}
\end{equation}

where the tilt angle $\theta_{tilt}$ is converted to an off-nadir pitch angle:

\begin{equation}
\theta_{pitch} = \theta_{tilt} -\frac{\pi}{2} 
\end{equation}

With the two camera matrices, $K$ and $R$, the procedure provided in Section~\ref{sec:p_matrix} was used to estimate the projection matrix.  

The alignment error between the two coordinate systems can be established by finding a rigid transform that aligns the estimated coordinate system to the ground truth coordinate system and measuring the error between the correspondence points.  A least-squares fitting of two 3-D point sets was used as defined in \cite{Arun}.  For the TownCentre dataset, the mean correspondence error was \SI{1.18}{\metre} with a standard deviation of $0.67$ using an average person height of \SI{1.8288}{\metre}.   An illustration of the correspondence points is shown in Figure \ref{f:img_correspondence_pts}.  The estimated ground plane view is shown in \ref{f:auto_correspondence_pts}.  Finally, the aligned systems are shown in \ref{f:gt_correspondence_pts}.   

\begin{figure}[h!]
\centering
\begin{subfigure}[h]{0.45\textwidth}
\centering
\includegraphics[height=1.5in]{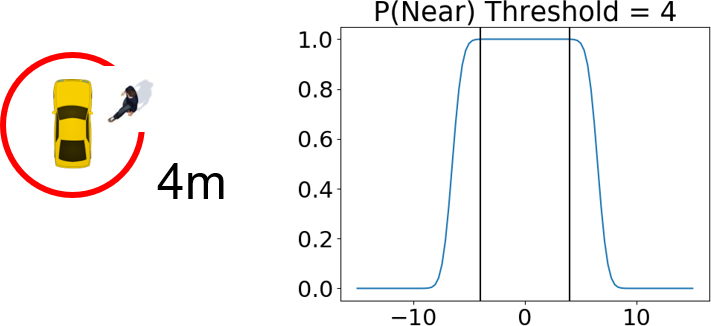}
\caption{$P(Near)$ with $\tau=\SI{4}{\metre}$}
\label{f:p_near}
\end{subfigure}%
\hfill
\begin{subfigure}[h]{0.45\textwidth}
\centering
\includegraphics[height=1.75in]{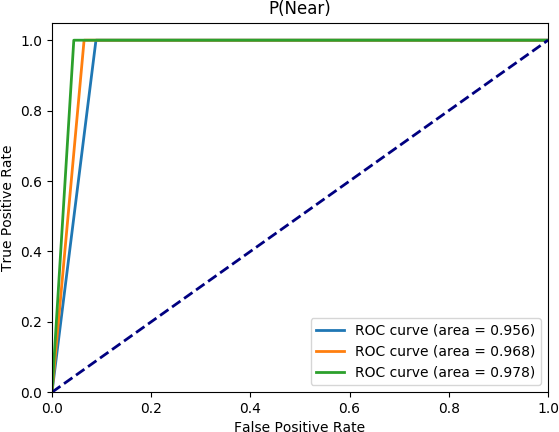}
\caption{ROC curves for average person heights of \SI{1.5748}{\metre} (blue), \SI{1.7018}{\metre} (orange), and \SI{1.8288}{\metre} (green)}
\label{f:p_near_roc}
\end{subfigure}
\caption{Vehicle proximity estimates}
\label{f:vehicle_proximity_estimates}
\end{figure}

\subsection{Spatial Proximity}
\label{sec:spatial_proximity}

If an activity detector is developed using a set of space-time predicates, a confidence score can be defined for the activity by establishing plausible probability distribution functions for each of the individual space-time predicates.  For instance, the activity \textit{Closing Trunk} might be defined when a person is facing the back of the vehicle, the person is near the vehicle, and the person's arm is moving down.   The probability of \textit{Closing Trunk} could then be expressed as:
 
\begin{equation}
P(Closing\ Trunk) = P(Facing) \cdot P(Near) \cdot P(Arm\ Moving\ Down) 
\end{equation}

Note that false alarms can be triggered with this simple definition.  For example, a person can be near the vehicle, facing the trunk, and waving to another person.  When their arm moves down after waving, $P(Closing\ Trunk) > 0$.

Nevertheless, this paper is concerned with $P(near)$, which is a function of the Euclidean distance between a person and a vehicle.  It can be designed similar to an activation function found in neural networks using the $erf(z)$ function and specifying a threshold $\tau$ as shown in \ref{f:p_near}.  In practice, a threshold distance of \SI{4}{\metre} is a good estimate to detect activities that involve a person interacting with a vehicle.    

The validity of this probability distribution function can be measured with a receiver operating characteristic (ROC) curve.  To populate the ROC curves, the Euclidean distances were measured between the people and the vehicles in the ground truth reference frame, using both the ground truth projection matrix and the aligned estimated projection matrix, to establish the True Positive and False Positive candidate sets.  If the ground truth distance was less than some threshold, the estimated distance was used to produce a True Positive score, otherwise it was use to produce a False Positive score.

With $\tau=\SI{4}{\metre}$, the ROC curves were established for three different projection matrix estimates using average person heights of \SI{1.5748}{\metre}, \SI{1.7018}{\metre}, and \SI{1.8288}{\metre}.  All heights produce ROC curves that prove this method would work reasonably well in practice, with the best area under the curve (AUC) of 0.98 for an average height of \SI{1.7018}{\metre}, as shown in \ref{f:p_near_roc}.

\begin{figure}[h!]
\centering
\begin{subfigure}[h]{0.45\textwidth}
\centering
\includegraphics[height=1.75in]{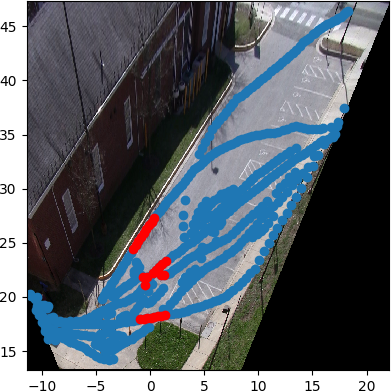}
\caption{VIRAT scene 0000 autocalibration result with inliers person footprints in red and outliers in blue}
\label{f:scene_0000_autocalib}
\end{subfigure}%
\hfill
\begin{subfigure}[h]{0.45\textwidth}
\centering
\includegraphics[height=1.75in]{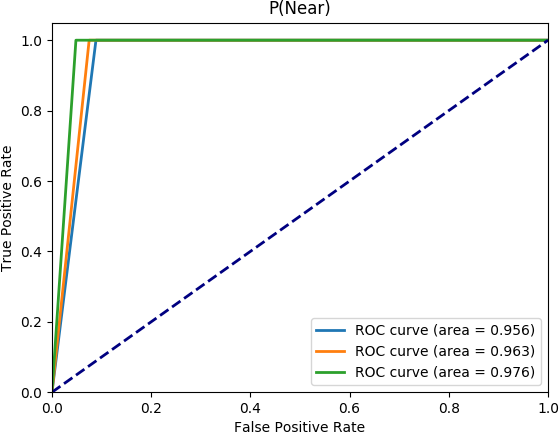}
\caption{ROC curves for average person heights of \SI{1.5748}{\metre} (blue), \SI{1.7018}{\metre} (orange), and \SI{1.8288}{\metre} (green)}
\label{f:p_near_roc_virat}
\end{subfigure}
\caption{VIRAT scene 0000 vehicle proximity estimates}
\label{f:virat_vehicle_proximity_estimates}
\end{figure}

To further test this method, scene $0000$ from the VIRAT dataset \cite{Oh_alarge-scale} was chosen, since ground truth homographies are provided by Kitware \cite{virat}.  The peak values from 500 Bird's Eye camera parameters were used to initialize the autocalibration process.  The mean correspondence error was \SI{2.1660}{\metre} with a standard deviation of $1.0772$,  using an average person height of \SI{1.8288}{\metre}.  The ROC curves for the same three heights used previously are shown in Figure \ref{f:virat_vehicle_proximity_estimates}, where an average height of \SI{1.8288}{\metre} provides the best result with an AUC of 0.98.

The $P(near)$ function is stable even though the Bird's Eye network predicted larger field of views than is represented by the scene images.  Additionally, the points on the head were not explicitly defined nor was the plum line to the ground from the top of the head.  The only points taken from the person detections were the centers of the top and bottom of the bounding boxes.

\section{Conclusion}
\label{sec:conclusion}

This paper has shown that a good spatial proximity estimate can be established between people and vehicles using a fully automated estimate of the 3-D scene geometry derived from a single frame camera pose estimator and pedestrian observations.  It is robust to deviations in the true camera pose and variations in persons heights.

In the future, this work can be extended to use tracks of people under the assumption that their height does not change as they walk through the scene.  Additionally, the vertical vanishing point can be used to dynamically adjust the person's pixel height estimate as a function of their spatial position in the image.  These enhancements will improve the accuracy of the camera height estimate and be useful for applications that require better accuracy of the 3-D locations of objects in the scene.

\bibliography{vsi_scene_alignment}
\bibliographystyle{plain}

\end{document}